\documentclass{article}

    \PassOptionsToPackage{numbers, compress}{natbib}

\usepackage[final]{neurips_2021}
\usepackage{graphicx}
\usepackage{subfig}

\usepackage[utf8]{inputenc} 
\usepackage[T1]{fontenc}    
\usepackage{url}            
\usepackage{booktabs}       
\usepackage{amsfonts}       
\usepackage{nicefrac}       
\usepackage{microtype}      
\usepackage[linesnumbered, ruled, vlined]{algorithm2e}
\usepackage{algorithmic}
\usepackage{pifont}
\usepackage[table,dvipsnames]{xcolor}
\title{Model-free feature selection to facilitate automatic discovery of divergent subgroups in tabular data}

\author{%
 Girmaw Abebe Tadesse \\
 IBM Research Africa \\
 \texttt{girmaw.abebe.tadesse@ibm.com } \\
 \And
 William Ogallo\\
 IBM Research Africa \\
 \texttt{william.ogallo@ibm.com } \\
 \And 
 Celia Cintas \\
  IBM Research Africa \\
 \texttt{celia.cintas@ibm.com} \\
\And
 Skyler Speakman \\
 IBM Research Africa \\
 \texttt{skyler@ke.ibm.com}
}

\begin{document}

\maketitle

\begin{abstract}
Data-centric AI encourages the need of cleaning, valuating and understanding of data in order to achieve trustworthy AI. Existing technologies, such as AutoML, make it easier to design and train models automatically, but there is a lack of a similar level of capabilities to extract data-centric insights. Manual stratification of tabular data  per a given feature of interest (e.g., gender) is limited to scale up for higher feature dimension, which could be addressed using  automatic discovery of divergent/anomalous subgroups. Nonetheless, these automatic discovery techniques often search across potentially exponential combinations of features that could be simplified using a preceding feature selection step.  Existing feature selection techniques for tabular data often involve fitting  a particular model (e.g., XGBoost) in order to select important features. However, such model-based selection is prone to model-bias and spurious correlations in addition to requiring extra resource to design, fine-tune and train a model.  In this paper, we propose a model-free and sparsity-based automatic feature selection (SAFS) framework to facilitate automatic discovery of divergent subgroups. Different from filter-based selection techniques, we exploit the sparsity of objective measures among feature values to rank and select features. We validated SAFS across two publicly available datasets (MIMIC-III and Allstate Claims) and compared  it with six existing feature selection methods. SAFS achieves a reduction of feature selection time by a factor of $81\times$ and  $104\times$, averaged cross the existing methods in the MIMIC-III and Claims datasets respectively. SAFS-selected features are also  shown to achieve competitive detection performance, e.g., $18.3\%$ of features selected by SAFS in the Claims dataset detected divergent samples similar to those detected by using the whole features with a Jaccard similarity of  $0.95$  but with a $16\times$ reduction  in detection time.
\end{abstract}

\section{Introduction}
\label{sec:intro}
AI research has been focused on building  sophisticated models that aim to exploit large baseline datasets across different domains.  Existing technologies, such as automated machine learning  (AutoML)~\cite{he2021automl}, make building and training AI models easy while achieving competitive performance with the state-of-the-art models. However, progress in research and practices to extract data-centric perspectives have been relatively limited, and hence significant resource is still allocated to clean and analyze data prior to feeding it to a model. Recent studies also show that baseline datasets, such as ImageNet~\cite{krizhevsky2012imagenet} contain considerable misannotations~\cite{northcutt2021confident}.  Data-centric AI is a growing field of research that aims to clean, evaluate data, and extract insights that are crucial for AI researchers/practitioners, domain-experts, and policy makers.

Stratification of data is a common technique to understand deviations across  different values of a feature of interest. Examples include variations in COVID-19 cases across age groups~\cite{levin2020assessing} or variations of child mortality across Sub-Saharan African countries~\cite{mejia2019age}.  However, such manual stratification does not scale up to encode interactions among higher number of features. Furthermore, human-level exploration is limited as we prioritize some hypotheses while ignoring others, early stopping of the exploration upon finding the first ``significant'' pattern in the data, and a tendency to identify patterns in the data that are not actually there (i.e., Type-1 error). To this end, automatic discovery techniques that   1) scale stratification to a higher number of features, 2) are less reliant on humans to pose the questions as this transfers our bias, 3) prioritize detecting patterns with the \emph{most evidence}, and 4) guard against false discoveries are necessary. 

Existing divergent (also known as outlier or anomalous) subgroup detection techniques could be mainly categorized into \textit{reconstruction}, \textit{classification} and \textit{probabilistic} groups~\cite{ruff2021unifying}.  The well-known principal component analysis and autoencoders are examples of reconstruction-based methods that, first, transform the data (e.g., to a latent space) so that anomalousness could be detected from failing to reconstruct the data back from the transformed data~\cite{hawkins2002outlier}. Classification-based approaches, particularly one-class classification, are often employed due to the lack of examples representing anomalous cases~\cite{khan2014one}. Furthermore, the traditional probabilistic models have also been used to identify anomalous samples using estimation of the normal data probability distribution, e.g., Gaussian mixture models~\cite{roberts1994probabilistic} and Mahalanobis distance evaluation~\cite{laurikkala2000informal}. Moreover, there are purely distance-based  methods, such as k-nearest neighborhood~\cite{gu2019statistical}, that do not require a prior training phase nor data transformations. Of note is that most existing methods infer anomalousness by exploiting individual sample characteristics rather than group-based characteristics. To this end, researchers proposed techniques, such as pysubgroup~\cite{lemmerich2018pysubgroup}, slice-finder~\cite{chung2019slice} and  multi-dimensional subset scanning (MDSS)~\cite{neill2013fast},  that aim to identify subsets of anomalous samples by exploiting group-level characteristics. 
Application of divergent group detection is crucial across different domains that include  healthcare~\cite{ogallo2021detection,kim2021out,zhao2019deep}, cybersecurity~\cite{xin2018machine},  insurance and finance sectors~\cite{zheng2018generative}, and industrial monitoring~\cite{hundman2018detecting}. For example, in healthcare, deviations could be \textit{erratic data annotations, vulnerable groups, least risk group}, and \textit{heterogeneous treatment effects}.

However, most existing detection techniques use  the whole input feature set as their search space, which includes exponentially growing combinations of feature values. For example, if there are $M$ binary features, there are $2^M-1$ possible combinations of feature values that may characterize a subgroup. In addition to the extended requirement of computational resources for large $M$, the identified subgroup might also be less interpretable when too many features are used to describe it. To this end, feature selection techniques could be employed to select $K < M$ features  to reduce the computational cost associated with detecting the subgroup due to the reduced search space while maintaining the detection performance.

 Existing feature selection techniques could be categorized as supervised and unsupervised based on their use of ground-truth labels in the data. Examples of supervised feature selection techniques include filters, wrappers, and embedded techniques \cite{miao_2016_a,molina_2002_feature,wanjiru2021automated}. In contrast, auto-encoders and principal component analysis are examples of unsupervised techniques that reduce the feature dimension in the latent space \cite{kumar_2015_a}. Existing filter-based feature selection techniques~\cite{wanjiru2021automated,molina_2002_feature} employ objective measures, such as mutual information, to encode the association between each feature and the outcome of interest. 
Wrapper methods use specific machine learning models to evaluate and select features \cite{miao_2016_a}. They measure the usefulness of features by learning a stepwise linear classifier or regressor using a recursive selection method, such as forward selection or backward elimination, until a stopping criteria is reached. 
On the other hand, embedded techniques utilize the output of model-fitting (e.g., using XGBoost~\cite{lundberg2020local} or CatBoost~\cite{wanjiru2021automated}) upon which feature importance ranking is extracted~\cite{molina_2002_feature,miao_2016_a}. Embedded techniques might require hyper-parameter fine-tuning of the model as in ~\cite{wanjiru2021automated} or training with a default tree-based classifier as in ~\cite{lundberg2020local}. 

Generally, existing wrapper and embedded feature selection techniques  mainly optimize over aggregated prediction performance of a trained model~\cite{wanjiru2021automated,molina_2002_feature,miao_2016_a}, which results in extra computational overhead (due to model fitting), and the feature selection output is prone to  model hyper-parameters, class imbalance, and under-/over-fitting. Even though existing filter-based feature selection techniques do not require model training, they are also limited in exploring the object measure variations across different unique values of a particular feature.

In this paper, we propose a sparsity-based  automatic feature selection framework (SAFS) which employs normalized odds ratio as an objective measure to evaluate the association between a feature value and the target outcome. Similar to the state-of-the-art filter-based techniques, SAFS is \textit{model-free} as it does not require training a particular model. In addition, SAFS encodes deviations of associations among feature values with the target using Gini-based sparsity evaluation metric~\cite{hurley2009comparing}.  Generally, the contributions of this work are as follows: 1) a model-free feature selection technique tailored to encode systemic deviations among subsets in a tabular data using a combination of association (i.e., normalized odds ratio) and sparsity (i.e., Gini index) measures that satisfy corresponding requirements; 2) we validate the proposed framework using two publicly available datasets: the MIMIC-III (Medical Information Mart for Intensive Care) dataset\cite{johnson2016mimic} and the Allstate Claim Severity dataset~\cite{claimkaggle}; 3) we compare the proposed SAFS to multiple existing feature selection techniques that include a mutual information based filter~\cite{estevez2009normalized}, Wrappers~\cite{miao_2016_a}, and Embedded techniques such as XGBoost~\cite{chen2016xgboost}, CatBoost~\cite{hancock2020catboost}, Committee~\cite{wanjiru2021automated} and Shap~\cite{lundberg2020local}. The results show that the proposed SAFS outperforms the baselines in ranking the features with significantly reduced ranking time, i.e., $81\times$ and  $104\times$ in MIMIC-III (with $41$ features) and Claims (with $109$ features) datasets, respectively. We  employ multi-dimensional subset scanning~\cite{neill2013fast} to validate the detection of the subgroups. Results show that $18.3\%$ and $48.8\%$ features selected by SAFS from Claim and MIMIC-III datasets, respectively, achieved competitive detection performance compared with the whole feature sets, with a Jaccard similarity of $0.93$ and $0.95$,  in the identified divergent samples.

The organization of the paper is as follows. Section~\ref{sec:proposed} describes the details of the proposed feature selection framework. Section~\ref{sec:setup} presents the datasets used for validation and existing techniques selected for comparison.  Section~\ref{sec:results} describes the results from the ranking of features across methods and the detection performance of divergent subgroups. Finally, Section~\ref{sec:conclusion} concludes the paper.

\section{Proposed framework}
\label{sec:proposed}
\begin{figure}
    \centering

    \includegraphics[width=1\linewidth]{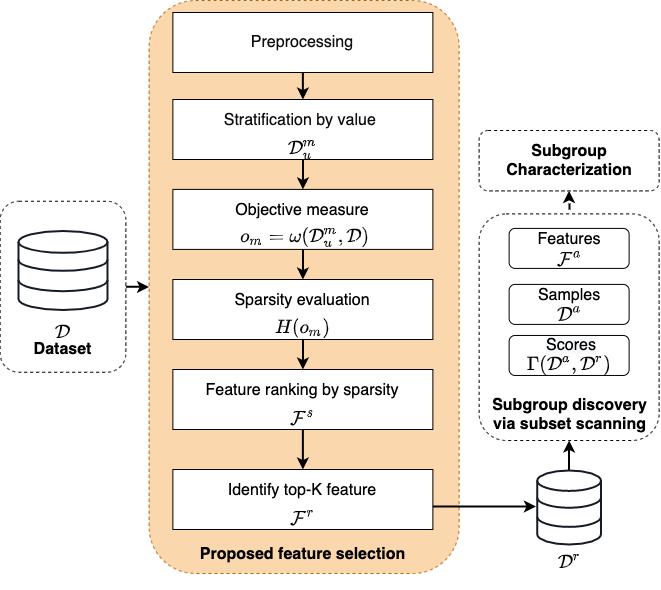}
    \caption{The overview of  proposed framework (SAFS) in which features are selected prior to automated divergence subgroup detection using sparsity-based evaluation of their associations with the outcome.}\label{fig:overview}

\end{figure}

The proposed framework is illustrated in Fig.~\ref{fig:overview} where the sparsity-based automatic feature selection (SAFS) precedes the divergent subgroup detection and characterization steps. SAFS exploits the sparsity of objective measures by first quantifying the association between each feature value and the outcome. Then a sparsity evaluation is applied to encode the deviations of the objective measures across unique values of a feature, and the features are ranked as per their sparsity values. The top$-k$ features in the SAFS rankings are then fed into an existing multi-dimensional subset scanning framework that automatically  stratifies and detects a divergent subgroup  with the extreme deviation from the expected. This group discovery process is followed by characterizing its feature descriptions, size, and divergent metrics between the identified subgroup and the remaining input data, such as odds ratio. Each component of the framework is described below in detail. 

\subsection{Problem Formulation}
Let $\mathcal{D}  = \{(x_i,y_i) | i = 1,2,\cdots, N\}$ denote a dataset containing $N$ samples where each sample $x_i$ is characterized by a set of $M$ discretized features  $\mathcal{F}=[f_1,f_2,\cdots,f_m, \cdots, f_M]$ and $y_i$ represents the binary outcome label. Note that each feature $f_m$ has $C_m$ unique values, $\hat{f}_m =\{\hat{f}_m^u\}_{u=1}^{C_m}$.  The proposed automated feature selection process is defined as a function $\mathcal{R}(\cdot)$ that takes $\mathcal{D}$ as input and  provides $\mathcal{D}^r$ represented by the top $K$ features, i.e., $\mathcal{D}^r=R(\mathcal{D},K) = \{(x^r_i,y_i) | i = 1,2,\cdots, N\}$ and $x^r_i$ is represented by  $\mathcal{F}^r =\{f^r_1,f^r_2,\cdots,f^r_k,\cdots,f^r_K\}$, where $K \leq M$.  Then an existing subgroup discovery method (MDSS)~\cite{neill2013fast}, $S(\cdot)$, takes $\mathcal{D}^r$  as input and identifies the anomalous subgroup ($\mathcal{D}^a$) represented by the set of anomalous features $\mathcal{F}^a=\{f^a_z\}_{z=1}^Z$, $Z \leq K \leq M$. 
The overall anomalous feature subspace is described as the logical (AND and OR) combinations of anomalous feature values as 
$\hat{\mathcal{F}}^a = \bigcap\limits_{z=1}^Z (\bigcup\limits_{h=1}^{H_z} \hat{f^{a}_{zh}}) $, where $\hat{f^{a}_{zh}}$ represents the $h^{th}$ value of the $f^{a}_z$ and $H_z < C_z$. Note that $\mathcal{F}^a \subseteq \mathcal{F}^r \subseteq  \mathcal{F}$. The detected divergent subgroup $\hat{\mathcal{D}}^a$ contains samples from $\mathcal{D}^r$ whose feature values are characterized by $\hat{\mathcal{F}}^a$, i.e.,  $\mathcal{D}^a=\{(x^a_j,y^a_j) | j=1,2,\cdots, P\}$, where $P < N$.
The divergence of the identified subgroup is evaluated using the anomalous score, $\Gamma(\mathcal{D}^a,\mathcal{D}^r)$ followed by post-discovery metrics such as odds ratio to quantify its divergence from the remaining subset of the input data.

\subsection{Automated Feature Selection}
 The sparsity-based automatic feature selection (SAFS) component in Fig.~\ref{fig:overview} is tasked with selecting the top $K$ features from a given $M$-dimensional feature space that are more useful for the subsequent anomalous subgroup discovery step. 
 Unlike the existing wrappers and embedded techniques in the state-of-the-art feature selection, SAFS is model-free and does not require training of a particular model. Similar to other filter-based techniques, SAFS employs a principled objective measure but to evaluate the  association between the outcome variable and each unique value of input feature.  To this end, SAFS uses Yule's Y-coefficient~\cite{yule1912methods}, which is a normalized odds ratio. We select  Yule's Y-coefficient as it is proven to be an objective measure that satisfies the most fundamental and additional proprieties for a good measure~\cite{tan2004selecting}.
 
 
 Given a feature $f_m$ with $C_m$ unique values, we manually stratify $\mathcal{D}$ per each feature value $\hat{f}_m^u \in \hat{f}_m$, resulting two subsets $\mathcal{D}_{m}^u$ and $\widetilde{\mathcal{D}_{m}^u}$, where $\mathcal{D} =\mathcal{D}_{m}^u \bigcup \widetilde{\mathcal{D}_{m}^u}$ and  $\mathcal{D}_{m}^u =\mathcal{D} | \hat{f}_m = \hat{f}_m^u$. For example, say a feature under-consideration is  ${f}_m = Sex$ with three unique values in $\mathcal{D}$: $\hat{f}_m^1=Female$, $\hat{f}_m^2= Male$ and $\hat{f}_m^3= Unknown/Missing$. Then stratifying for $\hat{f}_m^1$ gives $\mathcal{D}_{m}^1$ containing all samples in the $\mathcal{D}$ with $Sex = Female$ whereas $\widetilde{\mathcal{D}_{m}^1}$ containing the remaining samples in $\mathcal{D}$ with $Sex = Female$ or $Sex = Unknown/Missing$. In order to compute the Yule's Y coefficient, we generate a $2\times2$ contingency table from ${\mathcal{D}_{m}^u}$ and  $\widetilde{\mathcal{D}_{m}^u}$ as follows,

\begin{center}
 \begin{tabular}{l|c|c}
      & Y=1 &  Y=0  \\
\toprule
${\mathcal{D}_{m}^u}$ & $\alpha$ &  $\beta$  \\ \midrule
$\widetilde{\mathcal{D}_{m}^u}$ & $\delta$  & $\gamma$    \\

\bottomrule
\end{tabular}
\end{center}
where $\alpha$ is the number of samples in ${\mathcal{D}_{m}^u}$  with the binary outcome $Y=1$, $\beta$ is the number of samples in ${\mathcal{D}_{m}^u}$  with $Y=0$. Similarly, $\delta$ is the number of samples in $\widetilde{\mathcal{D}_{m}^u}$  with  $Y=1$, $\gamma$ is the number of samples in $\widetilde{\mathcal{D}_{m}^u}$  with $Y=0$. Note that $\alpha + \beta$ is size of ${\mathcal{D}_{m}^u}$ and $\delta + \gamma$ is the size of $\widetilde{\mathcal{D}_{m}^u}$, i.e.,  $N=\alpha + \beta + \delta + \gamma $.
Yule's Y objective measure for the feature value  $\hat{f}_m^u$ is computed as

 \begin{equation}
  o_m^u =\frac{\sqrt{P(\mathcal{D}_{m}^u,Y=1)P(\widetilde{\mathcal{D}_{m}^u},Y=0)}-\sqrt{P(\mathcal{D}_{m}^u,Y=0)P(\widetilde{\mathcal{D}_{m}^u},Y=1)}}{\sqrt{P(\mathcal{D}_{m}^u,Y=1)P(\widetilde{\mathcal{D}_{m}^u},Y=0)}+\sqrt{P(\mathcal{D}_{m}^u,Y=0)P(\widetilde{\mathcal{D}_{m}^u},Y=1)}} 
  \end{equation}
 \begin{equation}
 o_m^u =\frac{\sqrt{\alpha\gamma}-\sqrt{\beta\delta}} {\sqrt{\alpha\gamma}+\sqrt{\beta\delta}}
 \end{equation}
 
 Once $o_m^u$ is computed for $u=[1,2,\cdots,C_m]$ of feature $f_m$, we we employ Gini-index~\cite{hurley2009comparing} to evaluate the sparsity of Yule's Y coefficient across the feature values. Gini-index is selected as it is the only sparsity measure that satisfies the required six properties as described in ~\cite{hurley2009comparing}. These six properties include Dalton's four laws (\emph{Robin Hood, Scaling, Rising Tide and Cloning}) in addition to \emph{Bill Gates} and \emph{Babies}~\cite{hurley2009comparing}. Gini-index is computed over the ranked vector of objective measures $\overrightarrow{o_m} =[o_m^{(1)},o_m^{(2)},\cdots,o_m^{(C_m)}]$  in ascending order, i.e., $o_m^{(1)}\leq o_m^{(2)} \leq \cdots \leq o_m^{(C_m)}$, and $(1), (2), \cdots (C_m)$ are the indices after the sorting operation. Then the Gini-index ($\eta_m$)  of $f_m$ is then computed as
 \begin{equation}
     \eta_m=1-2\sum_{i=1}^{C_m} \frac{o_m^{(i)}}{||\overrightarrow{o_m}||_1}(\frac{C_m-i+\frac{1}{2}}{C_m})
 \end{equation}
where $||\cdot||_1$ represents the $l_1$ norm.  The Gini-index is computed for all the features $m=[1,2,\cdots, M]$ and they are ranked in decreasing order where a feature with the largest Gini-index takes the top spot. The summary of the steps for sparsity-based feature selection is shown in Algorithm~\ref{alg:algo_featureselection}. The follow up divergent subgroup detection step takes the top $K<M$ features as described below.

%

\begin{algorithm}[t]
\caption{Pseudo-code for the proposed automated feature selection (SAFS) based on the Gini sparsity of Yule's Y coefficients as objective measure between features and a binary outcome.}
\label{alg:algo_featureselection}

\SetKwFunction{IdentifyUniqueValues}{IdentifyUniqueValues}
\SetKwFunction{ZerosArray}{ZerosArray}
\SetKwFunction{SortDescendingIndices}{SortDescendingIndices}
\SetKwFunction{TopK}{TopK}
\SetKwFunction{YuleY}{YuleY}
\SetKwFunction{GiniSparsity}{GiniSparsity}
\SetKwFunction{Stratification}{Stratification}

\SetKwInOut{Input}{input}
\SetKwInOut{Output}{output}
\Input{Dataset: $\mathcal{D}$  $= \{(x_i,y_i) | i = 1,2,\cdots, N\}$,\\ Set of features: $\mathcal{F}$ $=[f_1,f_2,\cdots,f_m, \cdots, f_M]$,\\ Required number of features: $K$.}
\Output{Set of selected features: $\mathcal{F}^r$} 
\BlankLine
$\eta$ $\leftarrow$ \ZerosArray($M$)\;
\For{$f_m$ in $\mathcal{F}$}{
$\hat{f}_m$ $\leftarrow$ \IdentifyUniqueValues($f_m$)\;
$C_m$ $\leftarrow$ $|\hat{f}_m|$ \;
$o_{m}$ $\leftarrow$ \ZerosArray($C_m$)\;
\For{$u$ $\leftarrow$ $1~ \KwTo ~C_m$}{
 $\mathcal{D}^u_m$ $\leftarrow$ \Stratification($\mathcal{D}$,$\hat{f}_m^u$)\;
 $\widetilde{\mathcal{D}^u_m}$ $\leftarrow$  $\mathcal{D}^u_m$ - $\mathcal{D}^u_m$ \;
 $o_m^u$ $\leftarrow$ \YuleY($\mathcal{D}^u_m$,$\widetilde{\mathcal{D}^u_m}$)\;
 }

 $\eta_m$ $\leftarrow$ \GiniSparsity($o_m$)\;
 }
 
 $I$ $\leftarrow$ \SortDescendingIndices($\eta$)\;
 $\mathcal{F}^s$ $\leftarrow$ $\mathcal{F}[I]$\;
 $\mathcal{F}^r$ $\leftarrow$ \TopK($\mathcal{F}^s, K$)\;

\Return $\mathcal{F}^r$
\end{algorithm}

\subsection{Subgroup  Discovery  via Subset Scanning}
We employ Multi-Dimensional Subset Scanning (MDSS)~\cite{neill2013fast} from the anomalous pattern detection literature in order to identify significantly divergent subset of samples.  
MDSS could be posed as a search problem over possible subsets in a multi-dimensional array to identify the subsets with a systematic deviation between observed outcomes (i.e., $y_i$) and expectation of the outcomes, the latter of which could be set differently for variants of the algorithm. In the simple automatic stratification setting, the expectation is the global outcome average $\mu_g =\frac{\sum_{i=1}^N y_i}{N}$ in $\mathcal{D}^r$.
The deviation between the expectation and observation is evaluated by maximizing a Bernoulli likelihood ratio scoring statistic for a binary outcome, $\Gamma(\cdot)$. The null hypothesis assumes that the likelihood of the observed outcome in each  subgroup $\mathcal{D}^s$ is similar to the expected, i.e., $H_0: odds(\mathcal{D}^s)=\frac{\mu_g}{1-\mu_g}$; while the alternative hypothesis assumes a constant multiplicative increase in the odds of the observed outcome in the anomalous or extremely divergent subgroup $\mathcal{D}^a$, i.e., $H_1: odds(\mathcal{D}^a)=q\frac{\mu_g}{1-\mu_g}$ where $q\neq1$ ($q>1$ for extremely over observed subgroup (e.g., high risk population); and $0<q<1$ for extremely under observed subgroup (e.g., low risk population). The  anomalous scoring function for a subgroup ($\mathcal{D}^s$) with reference $\mathcal{D}^r$ is formulated as, $\Gamma(\mathcal{D}^s,\mathcal{D}^r)$ and computed as:
 \begin{equation}
\Gamma(\mathcal{D}^s,\mathcal{D}^r) = \max_q log(q)\sum_{i\in S} y_i - N_S * log(1-\mu_g + q\mu_g),
 \end{equation}
 where $N_S$ is the number of samples in $\mathcal{D}^s$. 
Divergent subgroup identification is iterated until convergence to a local maximum is found, and the global maximum is subsequently optimized using multiple random restarts.  
The characterization of the identified anomalous subgroup ($\mathcal{D}^a$) includes quantifying the anomalousness score $\Gamma(\mathcal{D}^a,\mathcal{D}^r)$, the analysis of the anomalous features and their values $\hat{\mathcal{F}}^a$, the size of the subgroup $N_S$, the odds ratio between $\mathcal{D}^a$ and $\widetilde{\mathcal{D}^a}$ and $95\%$ Confidence Interval (CI) of the odds ratio, the significance tests quantified using empirical p-value, and the time elapsed to identify $\mathcal{D}^a$.


\section{Experimental Setup}\label{sec:setup}

\subsection{Datasets}\label{subsec:daasets}
We employ two publicly available tabular datasets to validate the proposed feature selection framework and identify divergent subgroups in these datasets. These datasets are the Medical Information Mart for Intensive Care (MIMIC-III)~\cite{johnson2016mimic} and All State Claim Severity dataset (Claim)~\cite{claimkaggle}. 
MIMIC-III is a freely accessible critical care dataset recording vital signs, medications, laboratory measurements, observations and notes charted by care providers, fluid balance, procedure codes, diagnostic codes, imaging reports, hospital length of stay, and survival data. 
We selected a study cohort of adult patients (16 years or older) who were admitted to the ICU for the first time. The length of stay was greater than a day, with no hospital readmissions, no surgical cases, and having at least one day one chart events. The final cohort consisted of $19,658$ rows of patients data. We constructed $M=41$ features ($15$ numerical and $26$ categorical) based on observations made on the first 24 hours of ICU admission. The numerical features are later discretized. 
We defined the target outcome as a binary indicator variable $y_i$ such that  $y_i=1$ for patients who died within 28 days of the onset of their ICU admission, and $y_i=0$ otherwise. 

The Claim dataset is released by Allstate, an US-based insurance company as a Kaggle challenge~\cite{claimkaggle} to predict the severity of the claims.  In our validation, we use  $185,000$ training claim examples  with $109$ anonymized categorical features in our validation, and the numeric \emph{loss} feature is used as the outcome of interest.  We also transform the outcome to a binary variable using the median loss as a threshold, i.e., loss values greater than equal to the median loss are set to $y_i=1$, and loss values less than the median are set to $y_i=0$.

\subsection{Existing methods selected for comparison}\label{subsec:existing_methods}
In order to compare our sparsity-based feature selection framework, we selected multiple existing methods in the state-of-the-art for selecting features from  tabular data. These methods are Filter, Wrapper, and Embedded methods~\cite{miao_2016_a,wanjiru2021automated,molina_2002_feature}.  

Filter-based methods exploit the statistical characteristics of input data to select features independent of any modeling algorithms~\cite{wanjiru2021automated}.  
In this study, we implemented a filter method based on mutual information gain ~\cite{molina_2002_feature,vergara2014review}. Features are ranked in decreasing order of their mutual information, and top-ranked features are assumed to be more important than low-ranked features.

Wrapper methods ~\cite{wanjiru2021automated,miao_2016_a, molina_2002_feature} measure the usefulness of features by learning a stepwise Ordinary Least Squares regression and dropping less significant features recursively until a stopping rule, such as the required top $K$ features, is reached. In this study, we implemented a wrapper method using recursive feature elimination~\cite{guyon2002gene}.

Embedded methods select features based on rankings from a model. We implemented this by employing two tree-based classifiers, i.e., CatBoost~\cite{dorogush2018catboost} and XGBoost~\cite{chen2016xgboost} and a Committee vote~\cite{wanjiru2021automated} from both.  Unlike wrapper method, embedded methods begin with the fitting of the tree-based model followed by ranking features based on their importance.  In the case of Committee-based selection, the importance score from each of the two tree-based models is normalized using min-max scaling separately. Then the average of these importance scores is computed to rank the features. The three embedded methods above require calibration of the model. We also experimented with Shap~\cite{lundberg2020local}-value based feature importance using XGBoost classifier but using the default setting without calibration (herein referred to as \emph{Fast-Shap}).

\subsection{Setup}\label{subsec:setup}
We set-up the subgroup discovery task as form of automatic data stratification use cases in the two datasets as follows. The subgroup in  MIMIC-III dataset refers to a subset of patients with highest death risk compared to the average population in that dataset ($\mu_g = 17.2\%$). On the other hand, the subgroup discovery task in the Claim dataset is formulated as identifying a subset of claims  with the highest severity compared to the $\mu_g=42.2\%$ of claims possess higher  or equal to the median loss. 

For each dataset, we conducted subgroup discovery using the top $K$ features selected by the different feature selection methods examined. Specifically, we experimented with top $K$ $\in$ \{5, 10, 15, 20, 25, 30, 35, 40, 41\} features for the MIMIC-III dataset, and top $K$ $\in$ \{10, 20, 30,  40, 50, 60, 70, 80, 90, 100, 109\} features for the Claim dataset. Note that $K=41$ and $K=109$ represent using the entire original feature set in MIMIC-III and Claim datasets, respectively. 

To compare the different feature selection methods, we measured the computation times elapsed to rank the features as the first performance metric. Furthermore, we explore the similarity of the feature ranking across these methods using rank-based overlap \cite{webber2010similarity}. In addition, the output of the subgroup discovery algorithm using top $K$ features were also evaluated to inspect the usefulness of the selected features to detect divergent subgroup. We also compared the amount of time elapsed to identify the subgroup across different top $K$ values to determine the amount of computation time saved by using the selected top $K$ features rather than the whole input feature set. Lastly, we used the Jaccard similarity is employed to evaluate the similarity of the anomalous  samples detected by  using the selected top $K$  and the whole features.  Note that all the experiments are conducted on a MacBook Pro with macOS Big Sur OS v11.6, 2.9 GHz Quad-Core Intel Core i7 (processor) and 16 GB 2133 MHz LPDDR3 (memory).

\section{Results and Discussion}~\label{sec:results}

Table~\ref{tab:elpased_times} presents the time duration in seconds (s) elapsed by each feature selection method to rank features in the two datasets used for validation. The results show that the proposed sparsity-based feature selection framework achieved the ranking with the  shortest duration: $70.5$ seconds in Claim and just $3.3$ seconds in MIMIC-III datasets, resulting an average reduction of feature selection time by a factor of $104\times$ (in Claim) and $81\times$ (in MIMIC-III) compared to the existing methods. As expected, the wrapper method took the longest duration as it involves recursive fitting of a model and back-ward elimination of features. Considerably, the embedded methods also elapsed considerable time, particularly in the larger dataset due to the need for hyper-parameter tuning before selection. The committee vote approach also took longer than any of the XGBoost- and CatBoost-based embedded methods as it tries to normalize and merge their separate rankings. 
\begin{table}[t]
\caption{Comparison of elapsed times (in seconds) by different feature selection methods for ranking given feature sets  across two validation datasets: Claim and  MIMIC-III . The Proposed method achieved the least elapsed time in both the datasets.}\label{tab:elpased_times}
\centering
\begin{tabular}{lrr}
\toprule
& \multicolumn{2}{c}{Datasets} \\
       Method &   Claim &  MIMIC-III \\
\midrule
  Wrapper\cite{miao_2016_a,wanjiru2021automated} & 16444.3 &  777.4 \\
Committee\cite{wanjiru2021automated} & 13565.6 &  652.7 \\
       XGBoost\cite{chen2016xgboost} & 12306.1 &  124.7 \\
      CatBoost\cite{hancock2020catboost} &  1259.5 &   21.4 \\
   Filter\cite{wanjiru2021automated} &   494.4 &   14.4 \\
Fast-Shap\cite{lundberg2020local} &   114.1 &   11.3 \\ \hline
    \bf Proposed &    \bf 70.5 &    \bf 3.3 \\
\bottomrule
\end{tabular}
\end{table}

\begin{figure}[t]
    \centering
     \subfloat[MIMIC-III dataset]{
    \includegraphics[width=0.7\linewidth]{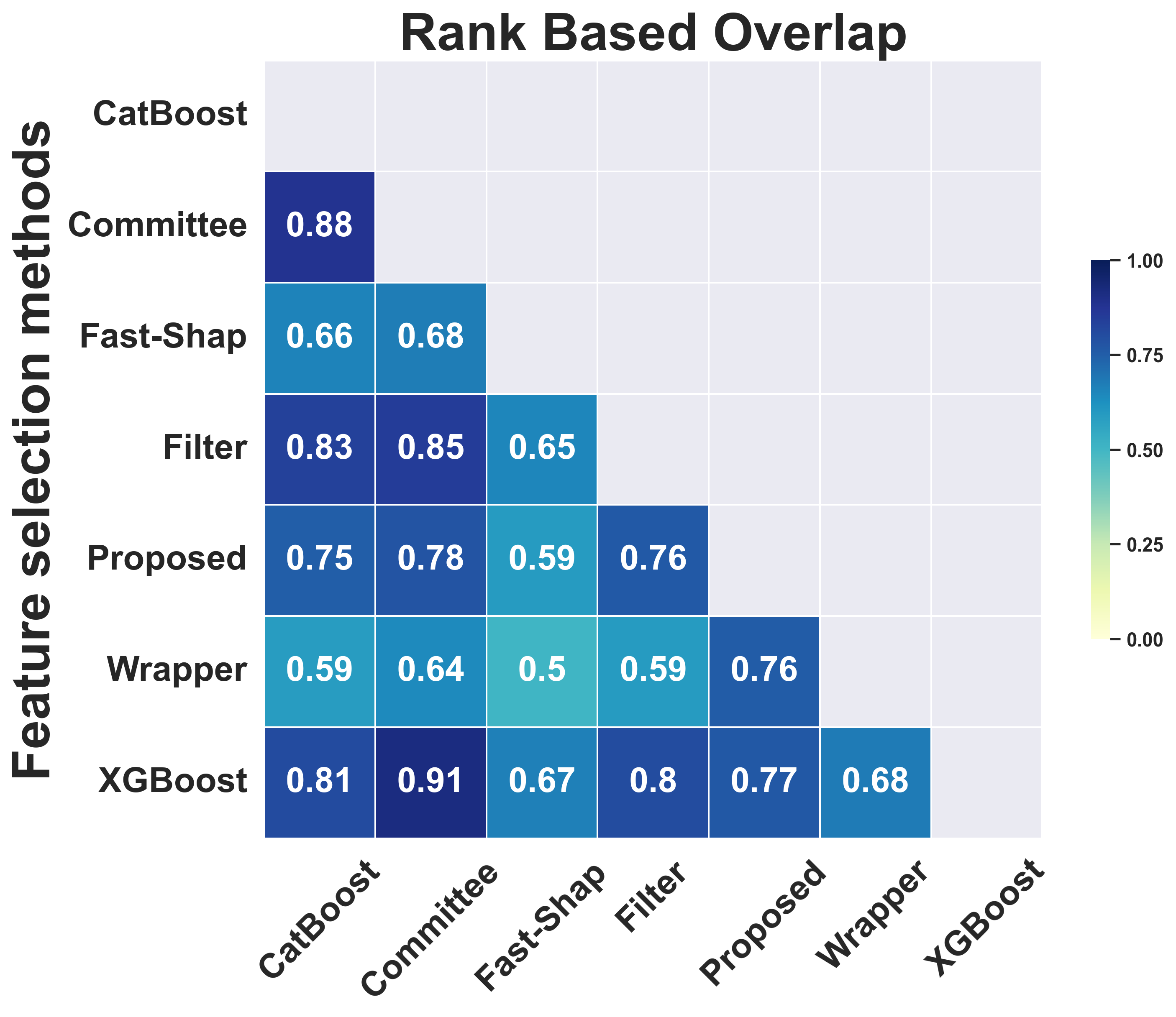}\label{fig:methods_overlap_mimc}}

            \subfloat[Claim dataset]{ \includegraphics[width=0.7\linewidth]{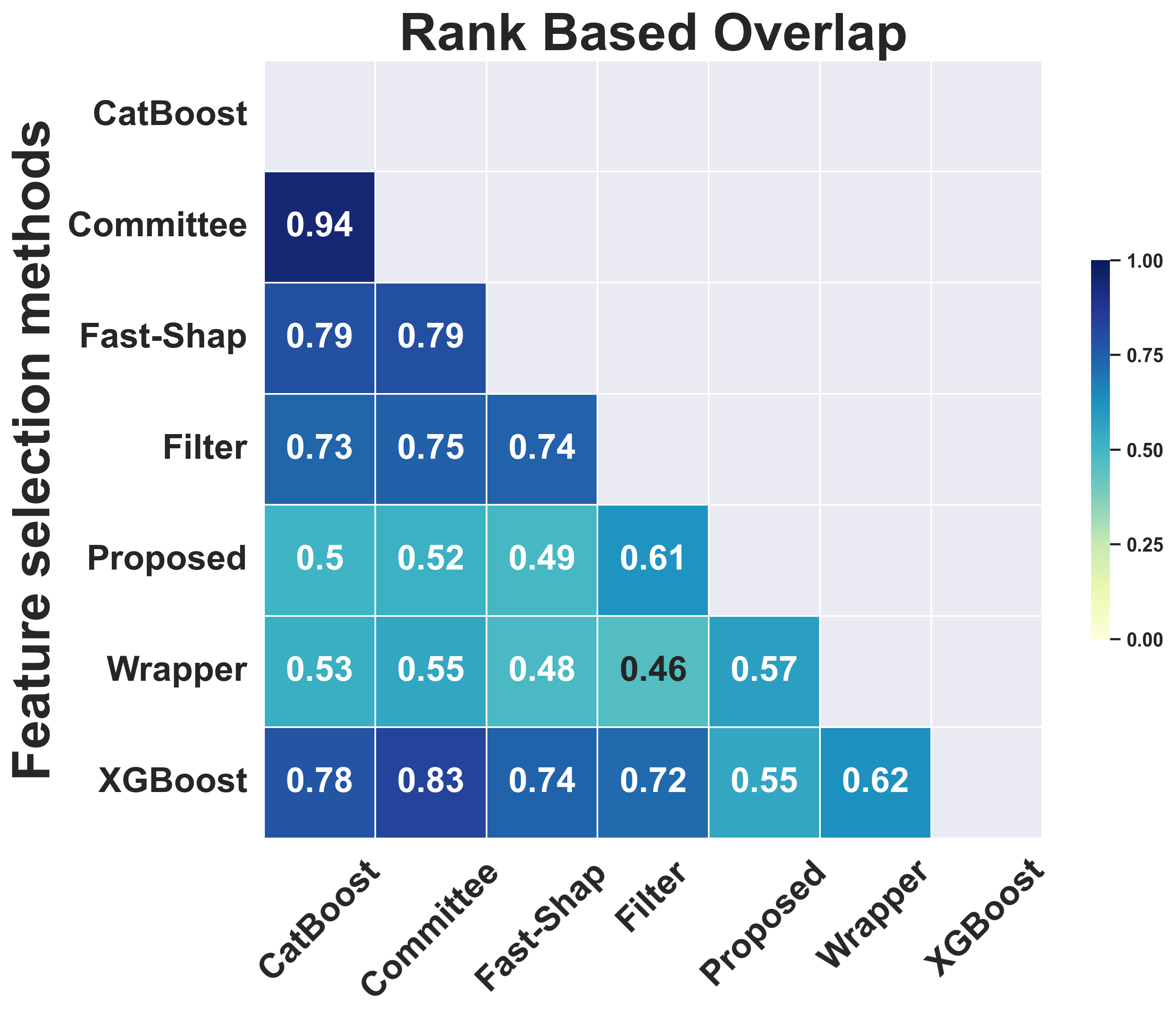}
       \label{fig:methods_overlap_claim}}
      \caption{Rank-based overlap scores quantifying the pairwise similarity of feature rankings by different selection methods.}\label{fig:methods_overlap}
\end{figure}

The pairwise similarity of rankings from different feature selection methods are illustrated in Fig.~\ref{fig:methods_overlap_mimc} for the MIMIC-III dataset and in Fig.~\ref{fig:methods_overlap_claim} for the Claim datasets using rank-based overlap scores. In both datasets, it is clear that the ranking from Committee vote achieved the higher similarity with CatBoost and XGBoost rankings. This is expected as the Committee vote is generated from the two rankings. The feature rankings from Fast-Shap, relatively, resemble the other embedded techniques, i.e., XGBoost and CatBoost. The overlapping of Fast-Shap and XGBoost are lower than the overlapping of CatBoost and XGBoost even though the only difference between Fast-Shap and XGBoost is that in the latter, the tuning of model parameters was conducted during fitting while Fast-Shap uses the default XGBoost parameters without tuning.  The proposed SAFS method outputs a higher similarity of ranking  with that of Filter method as both employ objective measures to quantify the association between features and outcome without training any model. The feature ranking from the Wrapper method are most different from the other methods suggesting that it is the least effective ranking method despite taking the longest duration to rank the features (see Table~\ref{tab:elpased_times}). 

\begin{figure}[t]
    \centering
        {
     \subfloat[MIMIC-III dataset]{
    \includegraphics[width=0.8\linewidth]{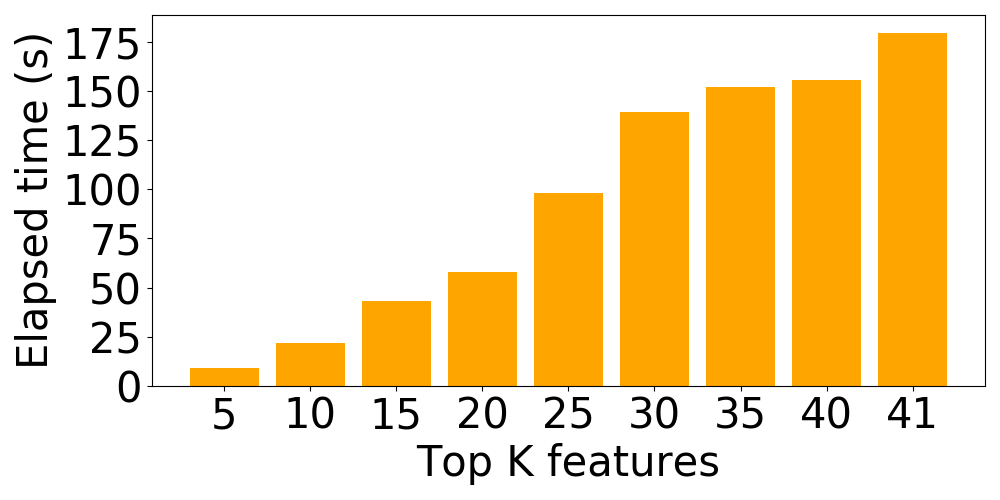}
   \label{fig:scanning_time_mimc}}
   
   \subfloat[Claim dataset]{
    \includegraphics[width=0.8\linewidth]{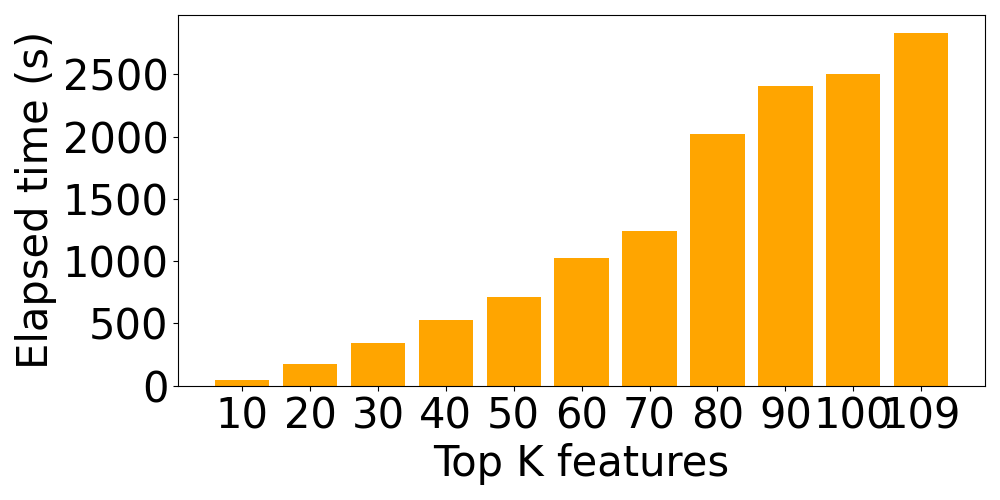}
    \label{fig:scanning_time_claim}}
 }
\caption{Comparison of the time taken to discover the divergent subgroup across different top $K$ features. Note $K=41$ and $K=109$ mean the whole feature set is used by the detection algorithm.}\label{fig:detection_times_k}
\end{figure}

\begin{figure}[t]
    \centering
      
\subfloat[MIMIC-III dataset]{
    \includegraphics[width=0.7\linewidth]{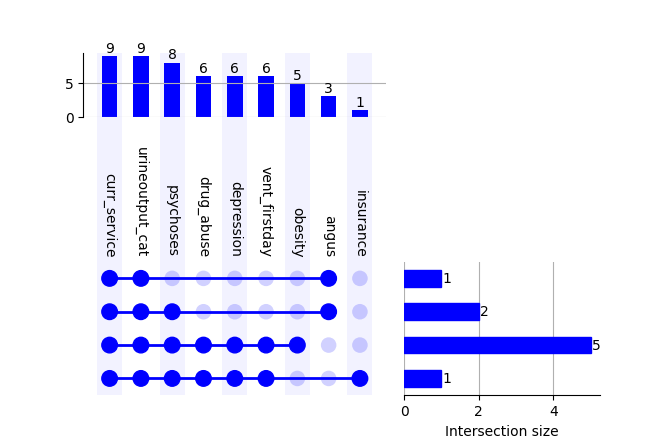}\label{fig:upset_mimic}}
    
    \subfloat[Claim dataset]{
    \includegraphics[width=0.7\linewidth]{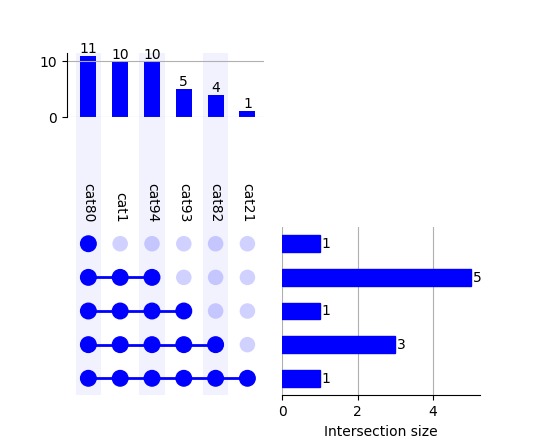}\label{fig:upset_claim}}
 \caption{Upset plots to visualize the similarity of identified anomalous features that  describe the detected divergent groups, across the two datasets (MIMIC-III and Allstate Claims), using different top $K$ features selected by the proposed SAFS.}\label{fig:upset_plots}
\end{figure}

Figure~\ref{fig:detection_times_k} shows a steep increase in the detection time (in seconds) with the increase in the number of features, thereby validating the plausibility of feature selection prior to the subgroup detection step. While it is clear that subgroup discovery is more efficient using a lower number of features selected in a principled way, it is important to evaluate the consistency of features identified to characterize the divergent subgroups detected. Ideally, similar descriptions of anomalous features should be achieved from scanning across different top $K$ features. To this end, we employed Upset plots to visualize the similarity of the features in the anomalous subsets identified by scanning over the top $K$ features using our proposed SAFS method as shown in Fig.~\ref{fig:upset_plots}. In MIMIC-III dataset (see Fig.~\ref{fig:upset_mimic}), two features \emph{curr\_service} and \emph{urineoutput\_cat} appears in the anomalous subset across all the different top $K$ values.
The feature \emph{psychoses} is also shown to appear in $8$ of $9$ different $K$ values. 
In the Claims dataset (Fig.~\ref{fig:upset_claim}), three features (\emph{cat80, cat1 and cat94}) are consistently detected as a subset of the anomalous features at least $10$ times out of $11$ different $K$ values. This validates the consistency of the identified group across different top $K$ values and reaffirms the benefit of using fewer selected features, which achieve similar subgroup but with significantly reduced computation time.

Table~\ref{tab:group_details_mimic_claim} presents a detailed comparison of the identified subgroups across different top $k$ features. The details include the number of features and the corresponding number of values describing the divergent subgroup, the number of samples in the subgroup, and percentage of the whole data in the subgroup. The table also shows the odds ratio (and its $95\%$ CI) of the outcome in the identified subgroup compared to the complement subset, and the significance of the divergence as measured by an empirical p-value.  Table~\ref{tab:claim} shows the detected subgroup details for the Claim dataset. It is clear that by just using 20 features ($18.3\%$ of the whole features), we identify a subgroup with $44181$ ($24\%$ of the whole data) characterized by the following logical combinations of \emph{features (and values)}:\emph{cat80 ('B' or 'C') and cat1 ('A') and cat94 ('B' or 'C' or 'D')}. The odds of experiencing severe claims is $8.79$ ($95\%$ CI: $8.58, 9.01$) higher in the identified subgroup than its complement. This identified subgroup is similar to the subgroup identified using  the whole $109$ features that results a subgroup of $4201$ samples ($23\%$ of whole data) and odds ratio of $9.30$ ($95\%$ CI: $9.07, 9.53$). The samples identified by using $K=20$ and $K=109$ features have a Jaccard similarity index of $0.95$. Table~\ref{tab:mimic} shows the same trend of achieving similar divergent subgroup using fewer number of features in the MIMIC-III dataset. Specifically, using just $20$ features ($48.8\%$ of  whole features), we identified subgroup size of $3183$ ($16\%$ of whole data) with a $3.86$($95\%$ CI: $3.54,4.20$) increase in odds of experiencing death within $28$ days of the onset of their ICU admission. This group is described with the following combination of features (and values): \emph{psychoses ('0')} and \emph{urineoutput\_cat ('1' or '2')} and \emph{drug\_abuse ('0')} and \emph{curr\_service ('0' or '5' or '6' or '8')} and \emph{depression ('0')}, \emph{insurance ('1' or '2' or '3' or  '4')} and \emph{vent\_firstday ('1')}.  This is similar to the group identified using the whole feature set ($K=41$), which results in a divergent subgroup of $3078$ ($16\%$ of whole data) with odds ratio of $3.95$ ($95\%$ CI: $3.63,4.31$), achieving a Jaccard similarity of $0.93$. 
The results also show that the number of features and their unique values to describe the identified subgroup unnecessarily grows with a larger number of top $K$ features. This validates the benefits of prior feature selection framework before subgroup detection to achieve better interpretability of the identified subgroup and significantly reduce the computation time. 

\paragraph{Limitations}: The proposed feature selection is shown to achieve competitively or better performance than existing feature selection techniques  as it does not require fitting of a particular model (compared to existing Wrappers and Embedded methods), and it exploits the variation of objective measures across the unique values of a feature (compared to existing Filter methods). However, its limitation is mainly rests on its requirement to discretize numeric features into bins.  
\begin{table}[t]
\caption{Post-discovery analysis of identified anomalous subgroups in the two datasets (Claim and MIMIC-III) per each top $K$ features selected for scanning using the MDSS-based automatic stratification. The details include the number of anomalous features detected (\emph{\#Feats}) and number of their unique values (\emph{\#Vals}), number of samples in the identified subgroup (\emph{Subset size}) and its percentage to the whole data (\emph{$\%$}. We also provide Jaccard similarity (\emph{Jacc. Sim.}) between identified samples from using top $K$ and the whole features,the odds ratio of outcome in the identified subset compared to the remaining data,  its $95\%$ confidence interval (\emph{CI})  and the empirical p-value (\emph{P}).  The results show that the discovery of the consistent anomalous subsets across top K values with slight decrease in the  subset size as the scanning with higher number of features resulting in more complex anomalous discovery.}\label{tab:group_details_mimic_claim}
    \centering
    
    {
\subfloat[Claim dataset]{
\begin{tabular}{p{0.5cm}p{1cm}p{1cm}cp{0.5cm}p{1cm}cc}

\toprule
K & \#Feats (\#Vals) & Subset size &   \% & Jacc. Sim. & Odds ratio & CI &  \textit{p}\\
\midrule
\midrule
    10 &         1 (1)  &          45707 &  25 &            0.81 &        8.52 &        (8.32, 8.72) &      0.0 \\
    20 &         3 (6)  &          44181 &  24 &            0.95 &        8.79 &        (8.58, 9.01) &      0.0 \\
    30 &         3 (6)  &          44181 &  24 &            0.95 &        8.79 &        (8.58, 9.01) &      0.0 \\
    40 &         3 (6)  &          44181 &  24 &            0.95 &        8.79 &        (8.58, 9.01) &      0.0 \\
    50 &         3 (6)  &          44181 &  24 &            0.95 &        8.79 &        (8.58, 9.01) &      0.0 \\
    60 &         3 (6)  &          44181 &  24 &            0.95 &        8.79 &        (8.58, 9.01) &      0.0 \\
    70 &         4 (9)  &          43708 &  24 &            0.95 &        8.91 &         (8.7, 9.13) &      0.0 \\
    80 &         5 (13) &          42333 &  23 &            0.99 &        9.25 &        (9.02, 9.48) &      0.0 \\
    90 &         5 (13) &          42333 &  23 &            0.99 &        9.25 &        (9.02, 9.48) &      0.0 \\
   100 &         5 (13) &          42333 &  23 &            0.99 &        9.25 &        (9.02, 9.48) &      0.0 \\
   109 &         6 (14) &          42101 &  23 &            1.00 &        9.30 &        (9.07, 9.53) &      0.0 \\

\bottomrule
\end{tabular}\label{table:claim_details}
}\label{tab:claim}
\vspace{0.5cm}
\subfloat[MIMIC-III dataset]{
\begin{tabular}{p{0.5cm}p{1cm}p{1cm}cp{0.5cm}p{1cm}cc}

\toprule
K & \#Feat (\#Vals) & Subset size &   \% & Jacc. Sim. & Odds ratio & CI &  \textit{p}\\
\midrule
     5 &         3 (8)  &           4577 &  23 &            0.32 &        3.60 &        (3.33, 3.89) &      0.0 \\
    10 &         4 (9)  &           4386 &  22 &            0.34 &        3.65 &        (3.38, 3.95) &      0.0 \\
    15 &         4 (9)  &           4386 &  22 &            0.34 &        3.65 &        (3.38, 3.95) &      0.0 \\
    20 &         7 (14) &           3183 &  16 &            0.93 &        3.86 &         (3.54, 4.2) &      0.0 \\
    25 &         7 (11) &           3078 &  16 &            1.00 &        3.95 &        (3.63, 4.31) &      0.0 \\
    30 &         7 (11) &           3078 &  16 &            1.00 &        3.95 &        (3.63, 4.31) &      0.0 \\
    35 &         7 (11) &           3078 &  16 &            1.00 &        3.95 &        (3.63, 4.31) &      0.0 \\
    40 &         7 (11) &           3078 &  16 &            1.00 &        3.95 &        (3.63, 4.31) &      0.0 \\
    41 &         7 (11) &           3078 &  16 &            1.00 &        3.95 &        (3.63, 4.31) &      0.0 \\

\bottomrule
\end{tabular}
}\label{tab:mimic}

}
\end{table}

\section{Conclusion and Future work}\label{sec:conclusion}
The model-centric approach has grown over the years to solve problems across different domains using sophisticated models trained on large datasets. 
While methods for data-centric insight extraction has not been given enough attention, they possess  bigger potential to understand, clean, and valuate data thereby improving the capability for more efficient performance using less resources. 
Automatic divergent subgroup detection could  be employed to answer different data-related questions, e.g., \emph{what are the high-risk population towards a particular disease?} Such automated techniques often do not require prior assignment of a feature of interest, and  they are scalable to high-dimensional feature input. However, detection of such subset of the data requires searching across potentially exponential combinations of input features that grow along with the number of input features. To this end, we propose a sparsity-based automated feature selection (SAFS) framework for divergent subgroup discovery that significantly reduces the search space and consequently, the amount of time required to complete the discovery and to improve the interpretation of the identified subgroups. SAFS employs a Yule's-Y coefficient as objective measure of effect between each feature value and an outcome; and then encodes the variations across values in a given feature using the Gini-index sparsity metric. Both Yule's-Y and Gini-index are  chosen as they were proven to satisfy the fundamental requirements of good objective measures and sparsity metrics, respectively. We validated our feature selection framework on two publicly available datasets: MIMIC-III (with $41$ features) and Allstate Claims (with $109$ features).  We also  compared SAFS with multiple existing feature selection techniques, such as Filters, Wrappers and Embedded techniques. Results showed that the proposed feature selection framework completed the ranking of features with the shortest duration in the two datasets, resulting an average reduction of feature selection time by a factor of $>92\times$ in the datasets (compared to the existing methods). Furthermore, the MDSS-based subgroup detection was employed to automatically identify divergent subgroups, in the two datasets. These subgroups refer to \emph{high death risk patients} in MIMIC-III and \emph{claims with high severity compared to the median loss} in Claims dataset. The detection results show the efficiency of our selection method that results in the discovery of similar divergent groups using just $\approx33\%$ of the original features (averaged across the datasets) compared to using the whole feature input.
%
%
Future work aims to improve the efficiency of SAFS and apply it across different use cases (e.g., heterogeneous treatment effects) and feature types (e.g., numeric features without discretization). Moreover, a similar approach could also be  employed to select nodes and layers in deep neural networks to detect adversarial/novel/outlier samples in latent spaces. 




\bibliographystyle{unsrt}
\bibliography{neurips_2021}

\end{document}